\documentclass{article}
\pdfoutput=1 

\usepackage[preprint]{neurips_2020}

\usepackage[utf8]{inputenc} 
\usepackage[T1]{fontenc}    
\usepackage{hyperref}       
\usepackage{url}            
\usepackage{booktabs}       
\usepackage{amsfonts}       
\usepackage{nicefrac}       
\usepackage{microtype}      
\usepackage{multirow}
\usepackage{graphicx}
\usepackage[accsupp]{axessibility}
\title{3rd Place Solution for NeurIPS 2021 Shifts Challenge: Vehicle Motion Prediction}

\author{Ching-Yu Tseng, Po-Shao Lin, Yu-Jia Liou, Kuan-Chih Huang
, Winston H. Hsu}

\begin{document}
\maketitle

\begin{abstract}

Shifts Challenge: Robustness and Uncertainty under Real-World Distributional Shift is a competition held by NeurIPS 2021. The objective of this competition is to search for methods to solve the motion prediction problem in cross-domain. In the real world dataset, It exists variance between input data distribution and ground-true data distribution, which is called the domain shift problem. In this report, we propose a new architecture inspired by state of the art papers. The main contribution is the backbone architecture with self-attention mechanism and predominant loss function. Subsequently, we won 3rd place as shown on the leaderboard.

\end{abstract}

\section{Introduction}
Prediction is one of the critical tasks in autonomous driving. For the purpose of generating commands to control the hardware of vehicles, we have to predict the appropriate trajectories of vehicles and control the vehicle to avoid collisions. As the ubiquity of the deep learning method, the prediction performance in a specific domain is promising. However, in the real world situation, the driving environments, weather conditions, and driver behaviors are extremely different from those in the different domains. Consequently, when the models are trained in the particular dataset, the model may not fit in other datasets. 

2021 Shifts Challenge focus on prediction task in shift-domains. Our target is to predict trajectories in 25 timestamps according to the given raster images. To tackle this practical problem, we propose a new architecture inspired by state-of-the-art papers. We modified the backbone model as NFNet to be our feature extractor because of its stability. We also add a self-attention layer since this architecture has achieved success in many time-related prediction tasks. Furthermore, we modified the loss function to achieve more robust performance. As a result, our method ranks 3rd place with 8.637 R-AUC CNLL in NeurIPS 2021 Shifts Challenge.

\section{Our Solution}
\begin{figure}
    \centering
    \includegraphics[scale=0.25]{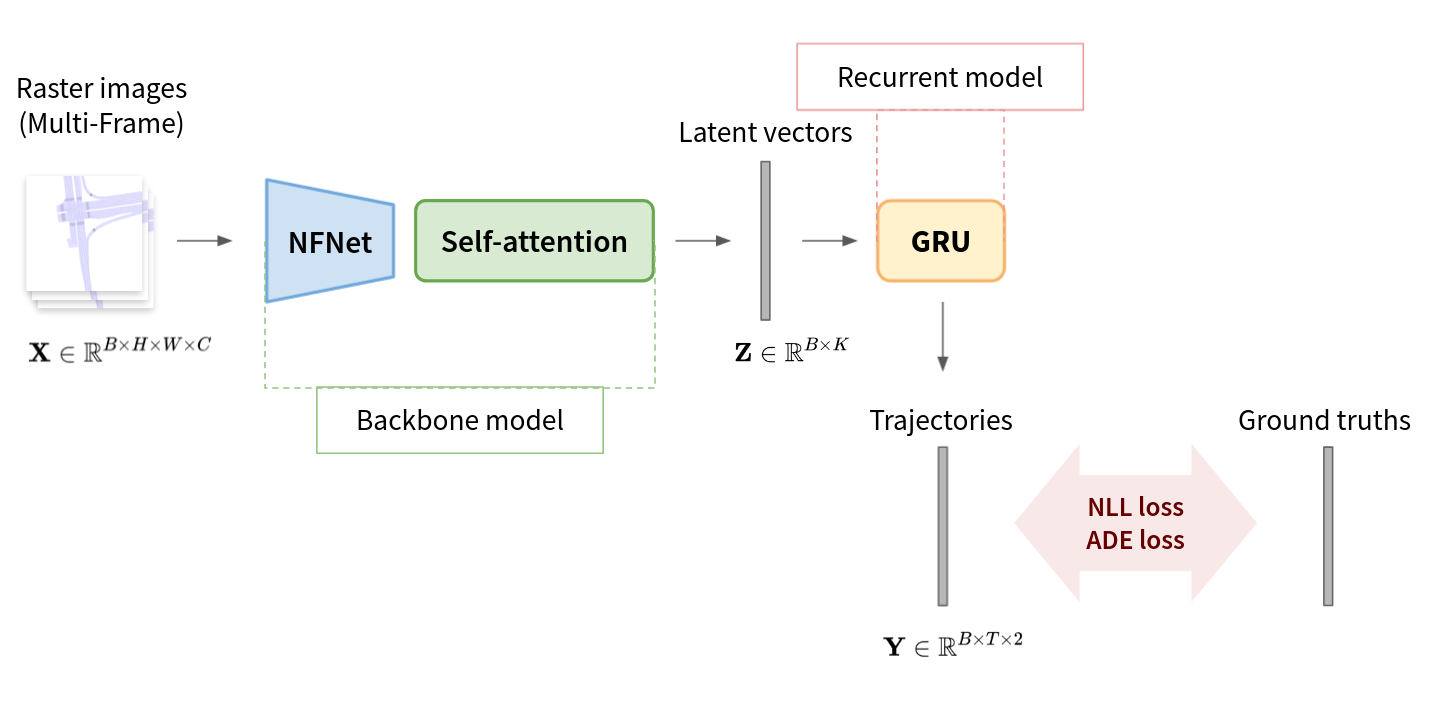}
    \caption{Base Model Architecture: To produce prediction according to multi-frame feature map, the baseline model first uses the backbone model to extract features and utilize recurrent model generate prediction according to latent vectors.}
    \label{fig:architecture}
\end{figure}
In this section, we detail our solution for this domain-shift problem by designing new model architectures. The domain-shift problem points to the situation that the training dataset and validation dataset come from different domains.

 Given the input raster images $\mathbf{X}$ which encode the first 5-second-information of vehicles, We aim to predict the last 5 seconds trajectories $\mathbf{Y}$ of objects. The information in raster images contains the conditions of dynamic objects(i.e., vehicles, pedestrians) which are described by their positions, orientations, linear accelerations, and velocities. 
 Our model is shown in Figure \ref{fig:architecture}. Our contributions mainly consists of 2 parts: (1) the enhancement for the new backbone model and feature extractor. (2) the revised loss function which leads to better performance.

\subsection{Baseline model}
The competition provides two baseline models\cite{bc, dim} and uses ensemble method\cite{rip} to enhance robustness. Both Behavior Cloning(BC)\cite{bc} and Deep Imitation Model(DIM)\cite{dim} adopt convolution backbone to compress information from raster images into a latent vector and apply the autoregressive model to generate tracks of vehicles according to latent vector. Behavior Cloning(BC)\cite{bc} assumes the autoregressive likelihood as single-variate gaussian distribution, Deep Imitation Model(DIM)\cite{dim} assumes multivariate normal distribution otherwise. 

After comparing the performance between BC\cite{bc} and DIM\cite{dim} methods, we choose BC\cite{bc} as our baseline architecture to improve due to superior performance. BC can be divided into 2 parts: The feature extraction backbone and the recurrent model.


\paragraph{Feature Extraction Backbone}
Given the input raster image $\mathbf{X}$, we use feature extraction backbone and self-attention layer(mentioned in next paragraph) $\mathit{f}$ to encode both spatial and temporal information of dynamic objects into latent embedding.
\begin{equation}
    \mathbf{Z} = \mathit{f}(\mathbf{X})
\end{equation}

Moreover, the baseline applies MobileNetV1\cite{mobilenets} as its backbone model. We have also replaced it with MobileNetV2\cite{mobilenetv2} and MobileNetV3\cite{mobilenetv3}, but all of them ended up with worse results. We assume the provided input data type is too simple to fit in most of the newest models due to model complexity. In the end, we find the NFNet\cite{nfnets} and select it to be our backbone model(feature extractor) due to its stability\cite{nfnets} of training.

\paragraph{Self-Attention Layer}
To further refine the features of the raster image, we add a self-attention layer\cite{attention}. The self-attention layer is the key mechanism in Transformer\cite{attention} which has a long-range dependency and takes global information into consideration. We divided the feature map into groups of pixels\cite{self-attention-block} and perform the self-attention to aggregate pixel-wise information.





\paragraph{Recurrent Model}
We choose GRU as our recurrent model which has a better performance compared with others.
With embedding from feature extraction as hidden states, the recurrent model makes predictions recursively. 

Given the embedding $\mathbf{Z}_{t}$ at timestamp t, with the output vector $\mathbf{Y}_0$ as zero vector, we use recurrent model $\mathit{g}$ to generate the predictions:

\begin{equation}
    \mathbf{Z}_{t} = \mathit{g}_{\rm{encoder}}(\mathbf{Y}_{t-1}, \mathbf{Z}_{t-1})
\end{equation}
\begin{equation}
    \mathbf{Y}_{t} = \mathit{g}_{\rm{decoder}}(\mathbf{Y}_{t-1}, \mathbf{Z}_{t})
\end{equation}
While $\mathbf{Y}_{t} \in \mathbb{R}^{\mathnormal{B} \times \mathnormal{T} \times 2}$ represents the location of a vehicle in 2D bird-eye-view map, $\mathbf{Z}_{t} \in \mathbb{R}^{\mathnormal{B} \times \mathnormal{K}}$ represents the hidden vector. $\mathnormal{B}$ and $\mathnormal{T}$ represents the dimensions of time and batch respectively.


    


\subsection{Loss Function}

In the beginning, we train our model with negative log-likelihood (NLL) loss like Eq. \ref{eq-4}. However, our model performs poorly on Average Distance Error(ADE) and Final Distance Error(FDE), thus we add the two metrics to minimize the distance between the predicted locations and ground-true locations. 
\begin{equation}
    {\rm{NLL}}(\mathbf{Y}) = -{\log(\mathnormal{p}(\mathbf{Y}))}
    \label{eq-4}
\end{equation}
\begin{equation}
        {\rm{Loss}} = \sum_{\mathbf{Y}} {-\log(\mathnormal{p}(\mathbf{Y};\theta))} + \sum_{l=1}^{\mathnormal{D}}(\hat{\mathbf{Y}}_l-\mathbf{Y}_l)^2 + (\hat{\mathbf{Y}}_f - \mathbf{Y}_f)^2
    \label{eq-5}
\end{equation}
 $\mathnormal{p}(\mathbf{Y};\theta)$ stands for the probability of predicted trajectory $\mathbf{Y}$ conditioned on model parameters $\theta$. $\mathbf{Y}_f$ stands for the final location of trajectory. In Equation \ref{eq-5}, the first part represents the original loss, the second one represents ADE Loss, and the last one represents FDE Loss.

\subsection{Ensemble Method}
To boost the performance, we follow the challenge setting to utilize Robust Imitative Planning(RIP)\cite{rip} to ensemble several models. 

\section{Experiments}
\subsection{Dataset and Evaluation}

\paragraph{Dataset}
We use the dataset provided by Yandex Self-Driving Group, which is the largest dataset for motion predictions so far. There are 27036 scenes in training sets and 9569 scenes in testing ones. Shifts Vehicle Motion Prediction dataset contains 600000 scenes from a different season, weather, location and times of a day. 
 This varied conditions suit the evaluation of the robustness. 

\paragraph{Evaluations metrics}
The organizers provide three kind of evaluation metrics: Average Distance Error(ADE), Final Distance Error(FDE), and Negative log-likelihood(NLL). Average Distance Error measures the Sum-Squared Errors between the predictions and ground-truth in each time step. Final Distance Error(FDE) only calculates the Sum-Squared Error of the last position between predicted and ground truth trajectories. The negative log-likelihood is the unlikelihood that predictions match grounds.

\subsection{Implementation Details}

After updating our model as mentioned in Sec.2.1, we train models on a single V100 machine for a day. The batch size is 512 and the learning rate is 1e-4. We resize the input feature map to 128 $\times$ 128. In the learning process, the AdamW\cite{wadam} optimizer and gradient clipping as 1.0 was utilized.

\subsection{Ablation Study and Comparison Results}



\paragraph{Ablation Study}

As shown in Table \ref{tab:table1}, we choose base models: DIM and BC as baselines. First, we change the backbones to EfficientNet, NFNet\cite{nfnets, efficientnet, efficientnetv2}, and the newest version of MobileNet\cite{mobilenetv2, mobilenetv3} to compare the different scales of parameters. The models with more parameters perform worse. 
We assume that it is due to the quantity of information from images mismatching the scale of parameters. That is, models with fewer parameters are sufficient to effectively extract the information of raster images.

Furthermore, we add a self-attention mechanism into our models and get a better result in Table \ref{tab:table1}. Finally, we add the loss of Average Distance Error(ADE) and Final Distance Error(FDE) to regulate the distance error, the results are verified in the Table \ref{tab:table1}. Although the DIM always give lowest Negative log-likelihood(NLL), it did not have a competitive result. 
Therefore, we give up the DIM to pursue performance.

\paragraph{Comparison Results}
After verifying the effectiveness of our designed base model, we use the aggregation model, RIP, to aggregate the predictions with the Worst Case Method(WCM), which samples several predictions per model and chooses the one with minimum confidence to promise the steadier results. As shown in Table \ref{tab:table2}, we suppressed the baselines and won the competitive results on the weighted sum of ADE and FDE. However, our result on MINADE and MINFDE did not perform well.
We end up in 3rd place.

\begin{table}[t]
\small
\centering
\begin{tabular}{|l|ccc|ccc|}
\hline


                                     &                & In Domain      &                  &                & Out of Domain  &                  \\
Method                               & ADE↓           & FDE↓           & NLL↓             & ADE↓           & FDE↓           & NLL↓             \\

\hline

 DIM + MobileNetV2(baseline)         & 2.450          & 5.592          & \textbf{-84.724} & 2.421          & 5.639          & \textbf{-85.134} \\
 BC + MobileNetV2(baseline)          & 1.632          & 3.379          & -42.980          & 1.519          & 3.230          & -46.887          \\
 BC + NFNet18                        & 1.225          & 2.670          & -53.149          & 1.300          & 2.893          & -53.130          \\
 BC + NFNet50                        & 1.360          & 2.963          & -50.605          & 1.392          & 3.066          & -51.317          \\
 BC + NFNet18 + Attention            & 1.174          & 2.549          & -56.199          & 1.325          & 2.852          & -54.476          \\
 BC + NFNet50 + Attention            & 1.155          & 2.504          & -56.291          & 1.265          & 2.770          & -54.730          \\
 BC + NFNet18 + ADE Loss             & 1.197          & 2.55           & -54.047          & 1.299          & 2.821          & -53.056          \\
 BC + NFNet18 + Attention + ADE Loss & \textbf{1.139} & \textbf{2.488} & -55.208          & \textbf{1.227} & \textbf{2.714} & -54.282          \\
     
\hline
\end{tabular}
\centering
\caption{Ablation Study on Shift Vehicle Motion Prediction Dataset}
\label{tab:table1}
\end{table}

\begin{table}[t]
\small
\centering
\begin{tabular}{|c|l|cccccc|}
\hline

Rank & Method           & \begin{tabular}[c]{@{}l@{}}Score\\ (R-AUC CNLL)\end{tabular} & CNLL↓           & WADE↓ & WFDE↓ & MINADE↓        & MINFDE↓        \\
\hline
- & baseline & 10.572         & 65.147 & 1.082          & 2.382          & 0.824 & 1.764 \\
1 & SBteam   & \textbf{2.571} & 15.676 & 1.850          & 4.433          & 0.526 & 1.016 \\
2    & Alexey \& Dmitry & 2.619                                                        & \textbf{15.599} & 1.326 & 3.158 & \textbf{0.495} & \textbf{0.936} \\
3 & Ours     & 8.637          & 61.864 & \textbf{1.017} & \textbf{2.264} & 0.799 & 1.719 \\

\hline
\end{tabular}
\centering
\caption{Quantitative Result of Top3 Final Submission: CNLL represents the weighted sum of NLL; WADE represents the weighted sum of ADE; WFDE represents the weighted sum of FDE;}
\label{tab:table2}
\end{table}

\section{Conclusion}
In the shift competition, we propose a novel architecture for the base model. Applying the ensemble method with our base model reaches  competitive performance. We have also implemented other state-of-the-art methods and compared the results with analysis. Moreover, we verify the robust performance of provided ensemble method. In the end, we win the third prize in the competition.



\small
\bibliographystyle{plain}
\bibliography{reference}

\end{document}